\documentclass[a4paper,9pt]{article}
\usepackage{amssymb,amsxtra,amsmath,ascmac}
\usepackage{graphicx}
\usepackage{color}

\usepackage{tabularx}
\usepackage{amsfonts}
\usepackage{latexsym}

\usepackage{times}
\usepackage{amssymb,amsxtra,amsmath,ascmac}

\newtheorem{theorem}{Theorem}
\newtheorem{definition}{Definition}

\newtheorem{lemma}{Lemma}

\newtheorem{example}{Example}
\newtheorem{remark}{Remark}

\title{
Singular leaning coefficients 
and efficiency in learning theory
}

\author{Miki Aoyagi\\
 \small
College of Science \& Technology, Nihon University, 
1--8--14, Surugadai, Kanda, Chiyoda-ku, \\
\hfill{\small Tokyo 101--8308 Japan. {(aoyagi.miki@nihon-u.ac.jp)}}
}

\date{}

\usepackage{amsmath}
\usepackage{amssymb}
\usepackage{amscd}

\usepackage{graphicx}
\usepackage{color}

\usepackage{lineno,hyperref}

\begin{document}
\maketitle


\noindent
{\bf Abstract} :

Singular learning models with non-positive Fisher information matrices include neural networks, reduced-rank regression, Boltzmann machines, normal mixture models, and others. These models have been widely used in the development of learning machines. However, theoretical analysis is still in its early stages.

In this paper, we examine learning coefficients, which indicate the general learning efficiency of deep linear learning models and three-layer neural network models with ReLU units. Finally, we extend the results to include the case of the Softmax function.

~


\section{Introduction}

{ Recently, deep neural networks have advanced significantly and have been applied to various types of real-world data. However, while many studies focus on numerical experiments, theoretical research has been relatively limited in comparison.
One reason for this is that deep neural networks are singular learning models, which cannot be analyzed using classical theories for regular models. Regular models refer to simple distributions, such as the normal distribution.

In this paper, we investigate the learning coefficients of deep neural networks.
The concept of learning coefficients originates from Bayesian machine learning and serves as a metric for evaluating model quality. These values are primarily used in information criteria for model selection methods.  
However, beyond their role in model selection, the behavior of learning coefficients also provides theoretical insights into the efficiency of learning models. For example, using these theoretical values~\cite{AoMulti}, deep neural networks  have been shown to explain the occurrence of double descent, a phenomenon where both generalization error and training error decrease simultaneously \cite{NKBYBS}.  
Additionally, recent studies have demonstrated that parameters with small learning coefficients tend to exhibit high stability during the learning process, making them valuable for theoretical research \cite{Furman2024EstimatingTL}.
In this paper, we extend the analysis of deep neural networks with linear units to those with nonlinear activation functions, specifically the ReLU function, demonstrating that this property also holds for networks with ReLU units.

\section{Bayesian learning theory}}


Assume that each sample \( (x_i,y_i)  \) is drawn from
a probability density function  
\(q(x,y)\)
and \((x,y)^n:=\{(x_i,y_i)\}_{i=1}^n\) are  \(n\) training
 samples selected  independently and identically
from \(q(x,y)\).
To estimate the true probability density function
\(q(x,y)\) using \((x,y)^n\) within the framework of Bayesian estimation, we consider  
 a learning model expressed in probabilistic form as \(p(x,y|w)\),
 along with an \textit{ a priori} probability density function  \(\varphi(w)\) 
 on a compact parameter set \(W\),
where \(w\in W \subset {\bf R}^d\) is a parameter. 
The \textit{ a posteriori} probability density function
\(p(w|(x,y)^n)\) is then given by:
$$p(w|(x,y)^n)=\frac{1}{Z_n({\beta})}\varphi(w)\prod_{i=1}^np(x_i,y_i|w)^{\beta},$$
where
$${Z_n}(\beta)=\int_W\varphi(w)\prod_{i=1}^np(x_i,y_i|w)^{\beta}{\rm d}w,$$
with $\beta$ representing an inverse temperature.

The Kullback-Leibler divergence
$D(p_1 \mid p_2)=\int p_1(z)\log \frac{p_1(z)}{p_2(z)}dz$
is a pseudo-distance between arbitrary probability density functions 
$p_1(z)$ and $p_2(z)$.

{ 
Define  
\[
L(w) = -E_{x,y}[\log p(x,y|w)] = \int q(x,y) \log \frac{q(x,y)}{p(x,y|w)} \,dxdy - \int q(x,y) \log q(x,y) \,dxdy
\]
\[
= D(q(x,y) \mid p(x,y|w)) - \int q(x,y) \log q(x,y) \,dxdy.
\]
Let \( w_0 \) be the optimal parameter that minimizes \( L(w) \) and, consequently, \( D(q(x,y) \mid p(x,y|w)) \) at \( w = w_0 \).  
Define the set of optimal parameters as  
\[
W_0 = \{w \in W \mid L(w) = \min_{w' \in W} L(w')\}.
\]}

Assume that its log likelihood function has relatively finite variance,
\[
{E}_{x,y}[ \log\frac{p(x,y|w_0)}{p(x,y|w)}]\geq c
{E}_{x,y}[( \log\frac{ p(x,y|w_0)}{p(x,y|w)})^2], \quad w_0\in W_0, w\in W,
\]
for a constant $c>0$.
{ 
Then, we have a unique probability density function  
$p_0(x,y) = p(x,y | w_0)$
for all $ w_0 \in W_0 $, 
meaning that the probability density function is the same for all 
$ w_0 \in W_0 $.}

Let 
$$f(x,y|w)=\log \frac{ p_0(x,y)}{p(x,y| w)}$$
and define its average error function as 
$$K(w)=E_{x,y}[f(x,y|w) ].$$
It is clear that  $K(w_0)=0$ for  all $w_0\in W_0$.

By applying Hironaka's Theorem \cite{Hi} to the function $K(w)$ at $w_0$, 
{ 
we obtain a proper analytic map $\pi$ from a manifold ${\cal U}$ }to a neighborhood of $w_0 \in W_0$:
\begin{equation}
\label{equation:k}
K(\pi(u))=u_1^{2k_1{ 
(u)}}u_2^{2k_2{ 
(u)}}\cdots u_d^{2k_d{ 
(u)}},
\end{equation}
where ${ 
u=}(u_1,\cdots,u_d)$ is a local analytic coordinate
system
on $U\subset {\cal U}$.
Furthermore, there exist analytic functions 
{ 
$a(x,y|u)$} and $b(u)\not=0$ such that:
\begin{equation}
f(x,y|\pi(u))=u_1^{{ 
k_1(u)}}u_2^{{ 
k_2(u)}}\cdots u_d^{{ 
k_d(u)}}{ 
a(x,y|u)},
\end{equation}
and
\begin{equation}
\label{equation:h}
\pi'(u) \varphi(\pi(u))=
u_1^{h_1{ 
(u)}}u_2^{h_2{ 
(u)}}\cdots u_d^{h_d{ 
(u)}}b(u).
\end{equation}

Let 
$$
\xi_n(u)=\frac{1}{\sqrt{n}}\sum_{i=1}^n\{u_1^{k_1{ 
(u)}}u_2^{k_2{ 
(u)}}\cdots u_d^{k_d{ 
(u)}}-{ 
a(x_i,y_i| u)}\}, 
$$
then,  we have 
an empirical process $K_n(\pi(u))$ such that
$$
nK_n(\pi(u))=\sum_{i=1}^nf(x_i,y_i| \pi(u))$$
$$=n u_1^{2k_1{ 
(u)}}u_2^{2k_2{ 
(u)}}\cdots u_d^{2k_d{ 
(u)}}-\sqrt{n}
u_1^{k_1{ 
(u)}}u_2^{k_2{ 
(u)}}\cdots u_d^{k_d{ 
(u)}}\xi_n(u).
$$

{ 
We introduce the learning coefficients
using $k_j(u)$ and $h_j(u)$, defined in \eqref{equation:k} and \eqref{equation:h},} 
as follows:
$$\lambda(w_0)=\min_{
{ U\subset {\cal U}}}
\min_{1\leq j\leq d}\frac{h_j{ (u)}+1}{2k_j{ (u)}},$$
and its order
$$\theta(w_0)= \max_{{ U\subset {\cal U}}}
{\rm Card}(\{j :  \frac{h_j{ (u)}+1}{2k_{ j(u)}}=\lambda(w_0)\}),$$
where
{  $U$ is a subset of ${\cal U}$ , $u$ is a local coordinate of $U$}
and 
${\rm Card}(S)$ denotes the cardinality of a set $S$.
Without loss of generality, we can assume that 
$$
\lambda(w_0)=\frac{h_1{ 
(u)}+1}{2k_1{ 
(u)}}=\frac{h_2{ 
(u)}+1}{2k_2{ 
(u)}}=\cdots=\frac{h_\theta{ 
(u)}+1}{2k_\theta{ 
(u)}}
<\frac{h_j{ 
(u)}+1}{2k_j{ 
(u)}}\, \, (\theta+1\leq j \leq d).
$$

In Bayesian estimation, the predictive probability density function of 
$(x,y)^n$
is given by:
$$p((x,y)^n)= {Z_n}(1)=\int \prod_{i=1}^n p(x_i,y_i | w) \varphi(w)dw.$$

According to Watanabe~\cite{Wa5}, for $w_0\in W_0$, we have
\begin{eqnarray*}
&&p((x,y)^n)\\
&=&  \prod_{i=1}^n p(x_i,y_i | w_0) 
\int \prod_{i=1}^n \frac{p(x_i,y_i | w)}{p(x_i,y_i | w_0)} \varphi(w)dw \\
&=&\prod_{i=1}^n p_0(x_i,y_i) 
\left(\frac{(\log n)^{\theta(w_0)-1}}{n^{\lambda(w_0)}}
{ \int} \int_0^\infty  \quad t^{\lambda(w_0)-1}\exp(- t+
\sqrt{t}\xi_n(u))dt du^*+o_p(\frac{(\log n)^{\theta(w_0)-1}}{n^{\lambda(w_0)}})\right),
\end{eqnarray*}
where
$\mu_j{ 
(u)}=-2\lambda k_j{ 
(u)}+h_j{ 
(u)}$,
$$du^*=\frac{\prod_{i=1}^\theta
\delta(u_i)\prod_{j=\theta+1}^d
u_j^{\mu_j{ 
(u)}}}{(\theta(w_0)-1)!\prod_{i=1}^\theta(2k_i{ 
(u)})}b(u)du,$$
and  $\delta(u)$ is Dirac's delta function.

This indicates that the most efficient parameter 
$w_0\in W_0$ 
  is the one with the smallest  $\lambda(w_0)$ and 
the largest $\theta(w_0)$,
where the Kullback-Leibler divergence between 
$q((x,y)^n)=\prod_{i=1}^n q(x_i,y_i ) $ and $p((x,y)^n)$
 is minimized:
\begin{eqnarray*}
&&D(q((x,y)^n) \mid p((x,y)^n) )= \int q((x,y)^n)  \log \dfrac{q((x,y)^n) }{p((x,y)^n)} \prod_{i=1}^n dx_idy_i \\
&=&\int q((x,y)^n) \log q((x,y)^n)  \prod_{i=1}^n dx_idy_i-\int q((x,y)^n) \log \prod_{i=1}^n p(x_i,y_i|w_0)  
\prod_{i=1}^n dx_idy_i  \\
&&+
  \int q((x,y)^n) \log \frac{\prod_{i=1}^n p(x_i,y_i|w_0) }{p((x,y)^n)} 
  \prod_{i=1}^n dx_idy_i\\
&=&\int q((x,y)^n) \log q((x,y)^n)  \prod_{i=1}^n dx_idy_i-\int q((x,y)^n) \log \prod_{i=1}^n p_0(x_i,y_i)  
\prod_{i=1}^n dx_idy_i  \\
&&+\lambda(w_0) \log (n)-(\theta(w_0)-1)\log\log (n)+O_p(1).  
 \end{eqnarray*}
Using these relations, we derive two model-selection methods: the "widely applicable Bayesian information criterion" (WBIC)~\cite{WBIC} 
and the "singular Bayesian information criterion" (sBIC)~\cite{Drton2}.

The learning coefficients are known as log canonical thresholds in algebraic geometry. Theoretically, their values can be obtained using Hironaka's Theorem. However, these thresholds have been studied primarily over the complex field or algebraically closed fields in algebraic geometry and algebraic analysis \cite{Kol, Mus, Kash}. There are significant differences between the real and complex fields. For instance, log canonical thresholds over the complex field are always less than one, while those over the real field are not necessarily so.
Obtaining these thresholds for learning models is challenging due to several factors, such as degeneration with respect to their Newton polyhedra and the non-isolation of singularities \cite{Ful}. As a result, determining these thresholds is a topic of interest across various disciplines, including mathematics.

Our purpose in this paper is to obtain $\lambda$ and $\theta$ for deep-layered linear neural networks,
and three-layer neural network models with ReLU units. Finally, we extend the results to include the case of the Softmax function.
In recent studies, we obtained 
exact values or bounded values of 
the learning coefficients for Vandermonde matrix-type singularities,
which are related to the three-layered neural networks and 
normal mixture models, among others~\cite{Ao2, Ao3, Ao8, ME4, FSDM}.
We have also exact values for the restricted Boltzmann machine~\cite{Ao7}.
Additionally, Rusakov and Geiger~\cite{RG2,RG} 
and Zwiernik \cite{Zw}, respectively,
obtained the learning coefficients for naive Bayesian networks and directed tree models with hidden variables.
Drton et~al. \cite{DLWZ} considered these coefficients 
for the Gaussian latent tree and forest models.
The paper \cite{Furman2024EstimatingTL} empirically developed a method to obtain the local learning coefficient for deep linear networks.

\section{Log canonical threshold}

\begin{definition}
\label{def:1}

Let ${ h}$ be an analytic function 
in neighborhood $W$ of 
 $w_0$,
and 
 $\varphi\geq 0$ be a $C^\infty$  function 
on $W$ that is also analytic in a neighborhood of $w_0$
with compact support.

Define the log canonical threshold 
$$\small c_{w_0}({ h},\varphi)=\sup\{c:|{ h}|^{-c}{ \varphi} \mbox{ is locally }L^2 \mbox{ in a neighborhood of  }w_0\}$$
over the complex field ${\bf C}$
and 
$$\small c_{w_0}({ h},\varphi)=\sup\{c:|{ h}|^{-c}{ \varphi} \mbox{ is locally }L^1 \mbox{ in a neighborhood of  }w_0\}$$
over the real field 
${\bf R}$.

\end{definition}

{ 
\begin{theorem}
\label{theorem:loglearn}
The learning coefficient $\lambda(w_0)$ is the log canonical threshold of the average error function over the real field.
\end{theorem}

\noindent
(Proof)

By applying Hironaka's Theorem \cite{Hi} to the function $h$ at $w_0$, 
we obtain a proper analytic map $\pi$ from a manifold ${\cal U}$ to a neighborhood of $w_0 \in W$:
\begin{eqnarray*}
h(\pi(u))&=&u_1^{\tilde{k}_1(u)}u_2^{\tilde{k}_2(u)}\cdots u_d^{\tilde{k}_d(u)},\\
\pi'(u) \varphi(\pi(u)) &=& u_1^{h_1(u)} u_2^{h_2(u)} \cdots u_d^{h_d(u)} b(u),
\end{eqnarray*}
where $u=(u_1,\cdots,u_d)$ is a local analytic coordinate
system
on $U\subset {\cal U}$.
Therefore,
$$\int |h|^{-c}\varphi dw=\sum_{U\subset {\cal U} }
\int_{U} |u_1^{\tilde{k}_1(u)}\cdots u_d^{\tilde{k}_d(u)}|^{-c}
u_1^{h_1(u)} \cdots u_d^{h_d(u)} b(u) du$$
$$=\sum_{U\subset {\cal U} }
\int_{U} |u_1^{-c\tilde{k}_1(u)+h_1(u)} \cdots u_d^{-c\tilde{k}_d(u)+h_d(u)}|b(u)du.$$

Since
$\int_{0}^1 x^{\alpha}dx<\infty$ if and only if $\alpha+1>0$,
we obtain
$$c_{w_0}(h,\varphi)=\min_
{U\subset {\cal U}}
\min_{1\leq j\leq d}\frac{h_j(u)+1}{\tilde{k}_j(u)}.$$
\hfill{Q.E.D.}

\begin{definition}  
Let us use the same notations as in the proof of Theorem \ref{theorem:loglearn}.  
We define \( \theta_{w_0}(h, \varphi) \) as the order of \( c_{w_0}(h, \varphi) \), given by  
\[
\theta_{w_0}(h, \varphi) = \max_{U\subset {\cal U}} {\rm Card}(\{j :  \frac{h_j(u)+1}{\tilde{k}_j(u)} = c_{w_0}(h, \varphi) \}).
\]  
For an ideal \( I \) generated by real analytic functions \( f_1,\dots,f_m \) in a neighborhood of \( w_0 \), we define  
\[
c_{w_0}(I,\varphi) = c_{w_0}(f_1^2+\cdots+f_m^2,\varphi),\quad 
\theta_{w_0}(I,\varphi) = \theta_{w_0}(f_1^2+\cdots+f_m^2,\varphi).
\]  
If $\psi(w^*)\not=0$, 
then denote
$c_{w_0}(h)=c_{w_0}(h, \varphi)$
and
$\theta_{w_0}(h)=\theta_{w_0}(h,\varphi)$
because the log canonical threshold and its order
are independent of $\varphi$.

\end{definition}  

\begin{theorem}  
If \( \varphi(w_0) \neq 0 \),  
then \( c_{w_0}(h, \varphi) \)  
and its order \( \theta_{w_0}(h, \varphi) \)  
are independent of \( \varphi \).  
\end{theorem}  

\noindent  
(Proof)  

If \( \varphi(w_0) \neq 0 \), then in a sufficiently small neighborhood  
\( V \) of \( w_0 \), there exist positive constants \( \alpha_1 \)  
and \( \alpha_2 \)  
such that  
\[
0 < \alpha_1 \leq \varphi(w) \leq \alpha_2.
\]  
Thus, we obtain  
\begin{eqnarray*}
&&\alpha_1 |h|^{-c}   \leq   |h|^{-c}   \varphi(w) \leq  \alpha_2 |h|^{-c}.
\end{eqnarray*}
This leads to  $c_{w_0}(h, \varphi) = c_{w_0}(h,1)$ and 
$\theta_{w_0}(h, \varphi) = \theta_{w_0}(h,1).$ 

\hfill{(Q.E.D.)}

}


Here,
{  $c_{w_0}(I,\varphi)$ 
and $\theta_{w_0}(I,\varphi)$}  for ideal $I$ is well-defined by Lemma \ref{lemma:Max}.

\begin{lemma}[\cite{Ao4}]
\label{lemma:Max}
Let $W$ be a neighborhood of $w^* \in{\bf R}^{d}$. 
Consider the ring of analytic functions on $U$.
Let ${\cal J}$ be the ideal generated by $f_1,\ldots,f_n$, which are analytic functions defined on $U$.
If $g_1,\ldots,g_m$ generate ideal ${{\cal J}}$, then 
$$c_{w^*}(f_1^2+\cdots+f_n^2,{ \varphi})=c_{w^*}(g_1^2+\cdots+g_m^2,{ \varphi}),
{ \theta_{w^*}(f_1^2+\cdots+f_n^2,\varphi)=\theta_{w^*}(g_1^2+\cdots+g_m^2,\varphi)}.$$
\end{lemma}

Define the norm of a matrix $C=(c_{ij})$
as $|| C ||=\sqrt{\sum_{i,j}|c_{ij}|^2}$.

\begin{definition}
For a matrix $C$, let $\langle C\rangle$ be the ideal generated by all { entries}
of $C$.

\end{definition}

\begin{theorem}
\label{theorem:k(w)}
Let $$h(x,w)=\sum_{0\leq i_1,\cdots,i_N\leq H} 
\tilde{h}_{i_1,\cdots,i_N}(w)x_1^{i_1}\cdots x_N^{i_N}$$
 be a polynomial with variables $x_1,\cdots,x_N$ and  $\tilde{h}_{i_1,\cdots,i_N}(w)$'s are continuous functions.
Let $q(x)$ be a continuous positive function on $X\subset {\bf R}^N$
with ${\rm Vol}(X)=\int_X q(x)dx>0$.
Then,  we have for some positive constants $\alpha_1,\alpha_2>0$ such that
\begin{eqnarray*}
&&\alpha_1\sum_{0\leq i_1,\cdots,i_N\leq H} 
\tilde{h}^2_{i_1,\cdots,i_N}(w)
\leq K(w)=\int_{X}h^2(x,w)q(x)dx
\leq \alpha_2\sum_{0\leq i_1,\cdots,i_N\leq H} \tilde{h}^2_{i_1,\cdots,i_N}(w).
\end{eqnarray*}

\end{theorem}

\noindent
{(Proof)}

Let ${\bf v}$ be the vector whose elements are $\tilde{h}_{0,0,\ldots,0}(w)$, $\tilde{h}_{1,0,\ldots,0}(w)$, $\tilde{h}_{0,1,\ldots,0}(w)$, 
$\ldots$, $\tilde{h}_{H,H,\ldots,H}(w)$. Let $C{ (x)}$ be the matrix whose $(i,j)$ elements are 
$x_1^{i_1}x_2^{i_2}\cdots x_N^{i_N} \cdot x_1^{j_1}x_2^{j_2}\cdots x_N^{j_N}$, 
where the $i$th element of ${\bf v}$ is $\tilde{h}_{i_1,\ldots,i_N}(w)$ 
and the $j$th element of ${\bf v}$ is $\tilde{h}_{j_1,\ldots,j_N}(w)$.

We have $h^2(x,w)={\bf v}^t C{ (x)}{\bf v}\geq 0,$
and $K(w)=\int_{X}h^2(x,w)q(x)dx
={\bf v}^t\int_{X}C{ (x)}q(x)dx  {\bf v}\geq 0. $
Also we have $K(w)=0$ if and only if ${\bf v}=0$,
 and  therefore $\int_{X}C{ (x)}q(x)dx $ is positive definite.
By setting $\alpha_1$ and $\alpha_2$ as the maximum and minimum eigenvalues of ${ \int_{X}C(x)q(x)dx} $, respectively, we complete the proof.
 
\hfill{(Q.E.D.)}

In the theorem above, we assume that \( h(x,w) \) is a polynomial in the variables \( x_1, \dots, x_N \).  
However, by leveraging the Noetherian ring property, this result can be extended to analytic functions \( h(x, w) \).

\section{Multiple neural network with linear units.}

In the paper~\cite{AoMulti},
the learning coefficients
for multiple-layered  neural networks
with linear units
were obtained.
In this section, we introduce the linear case with thresholds, which differs slightly from the case without thresholds.

We denote constants
 by superscript $*$, for example, $a^*$, $b^*$, and $w^*$.

Define matrices $A^{(s)}$  of size  $H^{(s)}\times H^{(s+1)}$
and vectors $B^{(s)}$ of  dimension $H^{(s)}$, for $s=1,\ldots,L$,
\begin{eqnarray*}
A^{(s)}=(a^{(s)}_{ij}),&&(1 \leq i \leq H^{(s)},1\leq j \leq H^{(s+1)}),\\
B^{(s)}=(b^{(s)}_{j}),&&(1 \leq i \leq H^{(s)}).
\end{eqnarray*}

Let $W$
be the set of parameters
$$W=\{ \{A^{(s)}\}_{1\leq s\leq L} , 
\{B^{(s)}\}_{1\leq s\leq L} \}.$$

Let $F^{(s)}(x)=A^{(s)}x+B^{(s)}$ be a function from ${\bf R}^{H^{(s+1)}}$ to
${\bf R}^{H^{(s)}}$.

Denote the input value by $x\in {\bf R}^{H^{(L+1)}}$
with probability density function $q(x)$
and output value $y\in {\bf R}^{H^{(1)}}$
for the multiple-layered neural network with linear units,
which is given by 
\begin{eqnarray*}
h(x,A,B)&=&F^{(1)}\circ F^{(2)}\circ \cdots \circ F^{(L)}(x)\\
&=&\prod_{s=1}^{L}A^{(s)}x+ \sum_{S=2}^{L}
\prod_{s=1}^{S-1}A^{(s)}B^{(S)}+B^{(1)}.
\end{eqnarray*}
Consider the  statistical model
with Gaussian noise,
$$p(y|x,w)=\frac{1}{(\sqrt{2\pi})^{H^{(1)}} }\exp
(-\frac{1}{2}||y-h(x,A,B)||^2), $$
$$p(x,y|w)=p(y|x,w)q(x).$$
The model has 
$H^{(1)}$ input units, $H^{(L+1)}$ output units,
and $H^{(s)}$ hidden units in each hidden layer.

Define the average log loss function
$L(w)$ by $L(w)=-E_{X,Y}[\log p(X,Y|w)]$
and assume the set of optimal parameters $W_0$ by 
\begin{eqnarray*}
&W_0=\{w_0\in W | L(w_0) =\min_{w'\in W} L(w')\}&\\
&=
\{ \{ A^{(s)}\}_{1\leq s\leq L} , 
\{B^{(s)}\}_{1\leq s\leq L}
\ | \  h(x,A,B)=h(x,A^*,B^*) \} .&
\end{eqnarray*}
We have
$p_0(x,y)=p(x,y|w_0)$ for all $w_0\in W_0$.

Moreover,
assume that 
the {\it a priori} probability density function 
$\varphi(w)$ is a $C^{\infty}-$ function with compact support $W$,
satisfying $\varphi(w^*)>0$.
{ Then, we have
  $K(w)=\frac{1}{2}\int_{X}(h(x,A,B)-h(x,A^*,B^*))^2q(x)dx$.

Let $r$ be the rank of $\prod_{s=1}^{L}A^{*(s)}$.}

\begin{definition}
\label{definition:MSa}

Let $r$ be a natural number
and $M^{(s)}=H^{(s)}-r$ for $s=1,\ldots,L+1$.
Define ${\cal M}\subset \{1,\ldots,L+1\}$ 
such that  
\begin{eqnarray*}
\small
\hspace*{-8mm}&&\ell ={\rm  Card}({\cal M})-1,\quad {\cal M}=\{S_1,\ldots,S_{\ell+1}\},\\
\hspace*{-8mm}&&M^{(S_j)}<M^{(s)} \mbox{ for } S_j\in  {\cal M} \mbox{ and } s\not\in {\cal M}, \\
\hspace*{-8mm}&&
\small
{\sum_{k=1}^{\ell+1}  M^{(S_k)} } \geq  \ell M^{(s)} \mbox{ for } s\in {\cal M},
\sum_{k=1}^{\ell+1} M^{(S_k)} <  \ell M^{(s)}\mbox{ for } s \not \in  {\cal M}.
\end{eqnarray*}
Let $M$ be the integer such that 
$M-1< \dfrac{\sum_{k=1}^{\ell+1} M^{(S_k)}}{\ell} \leq M,$
and 
$a=\sum_{k=1}^{\ell+1} M^{(S_k)}-(M-1)\ell.$

Let
\begin{eqnarray*}
&&\lambda(H^{(1)},H^{(2)},\cdots,H^{(L+1)},r)\\
&=&
\frac{-r^2+r(H^{(1)}+H^{(L+1)})}{2}
+\frac{a(\ell-a)}{4\ell}
+\frac{1}{4\ell}
{(\sum_{j=1}^{\ell+1} M^{(S_j)})^2 }
 -
\frac{1}{4} \sum_{j=1}^{\ell+1} (M^{(S_j)} )^2\\
&=&
\frac{-r^2+r(H^{(1)}+H^{(L+1)})}{2}
+\frac{Ma+(M-1)\sum_{j=1}^{\ell+1} M^{(S_j)} }{4} -
\frac{1}{4} \sum_{j=1}^{\ell+1} (M^{(S_j)} )^2
\end{eqnarray*}
and 
$\theta(H^{(1)},H^{(2)},\cdots,H^{(L+1)},r)=a(\ell-a)+1.$

\end{definition}

{ 

In simple terms, \( {\cal M} = \{ S_1, \ldots, S_{\ell+1} \} \)  
represents the set of relatively smaller values 
\( M^{(S_i)} \) within \( M^{(s)} \),  
since \( \lambda \) must be the minimum value among \( \frac{h_j+1}{2k_j} \).

}

\begin{remark}
If \  ${\rm Card }\{s \ | \  M^{(s)}=0\}\geq 1$,  then we have 
$\lambda(H^{(1)},H^{(2)},\cdots,H^{(L+1)},r)=
\frac{-r^2+r(H^{(1)}+H^{(L+1)})}{2}$ and 
$\theta(H^{(1)},H^{(2)},\cdots,H^{(L+1)},r)=1$.
\end{remark}

The log canonical threshold  $\lambda$ and its order $\theta$
are as follows.
\begin{theorem}
\label{theorem:part2}

We have
\begin{eqnarray*}
\lambda&=&
\frac{H^{(1)}}{2}
+\lambda(H^{(1)},H^{(2)},\cdots,H^{(L+1)},r)
\end{eqnarray*}
and 
$$\theta=
\theta(H^{(1)},H^{(2)},\cdots,H^{(L+1)},r).$$

\end{theorem}

\noindent
(Proof)

{ 
Let \( A \) be a matrix, and let \( B \) and \( x \) be vectors.  
Since  
$
\|Ax + B\|^2 = \sum_{i} \left( \sum_j a_{ij}x_j + b_i \right)^2,
$  
we have  
\[
\int \|Ax + B\|^2 q(x) dx  
= \int \sum_{i} \left( \sum_j a_{ij}x_j + b_i \right)^2 q(x) dx  
= \sum_{i} \int \left( \sum_j a_{ij}x_j + b_i \right)^2 q(x) dx.
\]  
By Theorem \ref{theorem:k(w)}, there exist positive constants \( \alpha_1 > 0 \) and \( \alpha_2 > 0 \) such that  
\[
\alpha_1 \left( \sum_{j} a_{ij}^2 +  b_i^2 \right)  
\leq \int \left( \sum_j a_{ij}x_j + b_i \right)^2 q(x) dx  
\leq \alpha_2 \left( \sum_{j} a_{ij}^2 +  b_i^2 \right).
\]  
Therefore, we obtain  
\[
c_{w^*} \left( \int \|Ax + B\|^2 q(x) \, dx \right)  
= c_{w^*} ( \|A\|^2 + \|B\|^2), \quad 
\theta_{w^*} \left( \int \|Ax + B\|^2 q(x) \, dx \right)  
= \theta_{w^*} ( \|A\|^2 + \|B\|^2).
\]  
This implies that, since  
\[
K(w) = \frac{1}{2} \int \Big\| 
\left(
\prod_{s=1}^{L} A^{(s)} - \prod_{s=1}^{L} A^{*(s)}
\right) x + \sum_{S=2}^{L}
\prod_{s=1}^{S-1} A^{(s)} B^{(S)} + B^{(1)}
- \sum_{S=2}^{L}
\prod_{s=1}^{S-1} A^{*(s)} B^{*(S)} - B^{*(1)}
\Big\|^2 q(x) \, dx,
\]  
we obtain  
\[
c_{w^*}(K(w)) =
c_{w^*} \Big( \big\| \prod_{s=1}^{L} A^{(s)} - \prod_{s=1}^{L} A^{*(s)} \big\|^2
+ \big\| \sum_{S=2}^{L}
\prod_{s=1}^{S-1} A^{(s)} B^{(S)} + B^{(1)}
- \sum_{S=2}^{L}
\prod_{s=1}^{S-1} A^{*(s)} B^{*(S)} - B^{*(1)}
 \big\|^2 \Big),
\]  
and similarly for \( \theta_{w^*}(K(w)) \).  }

Since we can change the variables by 
$$B'^{(1)}=\sum_{S=2}^{L}
\prod_{s=1}^{S-1}A^{(s)}B^{(S)}+B^{(1)},$$
we obtain the theorem from Lemma \ref{lemma:Max} (2) and the Main Theorem in the paper \cite{AoMulti}.

\hfill{(Q.E.D.)}

\section{Multiple-layered neural networks with 
activation function ReLU (Rectified Linear Unit function)}

In this paper, we consider the case in multiple-layered neural networks with 
activation function ReLU (Rectified Linear Unit function).

For a matrix $A=(a_{ij}), a_{ij}\in {\bf R}$, 
define $$A_{+}=(\max\{0,a_{ij}\}) .$$

{ 
Denote the input value by $x\in {\bf R}^{H^{(L+1)}}$
with probability density function $q(x)$
and output value $y\in {\bf R}^{H^{(1)}}$
for the multiple-layered neural network with ReLU units,
which is given by 
$
h_{+}(x,A,B)=F^{(1)}_+\circ F^{(2)}_+ \circ \cdots \circ F^{(L)}_+(x).
$
Consider the  statistical model with Gaussian noise,
$$p(y|x,w)=\frac{1}{(\sqrt{2\pi})^{H^{(1)}} }\exp
(-\frac{1}{2}||y-h_+(x,A,B)||^2), $$
$$p(x,y|w)=p(y|x,w)q(x).$$}

\begin{definition}
Let 
$$
V^{(L)} (F_{i_1}^{(L)},\ldots,F_{i_k}^{(L)})
=\{x \ | \  F_{i_1}^{(L)}(x)\geq  0,   \cdots, 
F_{i_k}^{(L)}(x)\geq  0, \quad F_{j}^{(L)}(x)<0 \  \mbox{\rm for }\   j\not =i_1,\ldots,i_k \}
$$
and
\begin{eqnarray*}
\Omega^{(L)}&=&\{ V \   | \ 
V= V^{(L)} (F_{i_1}^{(L)},\ldots,F_{i_k}^{(L)}) \mbox{ \rm 
for some }  1\leq {i_1}, {i_2},\cdots, {i_k}\leq H^{(L)} \ \}.
\end{eqnarray*}

Inductively, we define
\begin{eqnarray*}
V^{(s)} (F_{i_1}^{(s)},\ldots,F_{i_k}^{(s)})
&=&\{x\in V^{(s+1)}\in \Omega ^{(s+1)}  \ | \ 
  F_{i_1}^{(s)}\circ
F^{(s+1)}\circ  \cdots \circ F^{(L)}  (x)\geq  0,  \\
&&\qquad \ldots, 
F_{i_k}^{(s)}\circ F^{(s+1)}\circ  \cdots \circ F^{(L)}  (x)\geq  0 ,\\
&&
\qquad  F_{j}^{(s)}\circ F^{(s+1)}\circ \cdots \circ F^{(L)}  (x)< 0
 \  \mbox{\rm for }\   j\not =i_1,\ldots,i_k \}
\end{eqnarray*}
and
\begin{eqnarray*}
\Omega^{(s)}&=&\{ V \   | \ 
V= V^{(s)} (F_{i_1}^{(s)},\ldots,F_{i_k}^{(s)}) \mbox{ \rm 
for some }  1\leq {i_1}, {i_2},\cdots, {i_k}\leq H^{(s)} \ \}.
\end{eqnarray*}


\end{definition}

By the definition, we have 
$V_1\cap V_2= \emptyset $
for $V_1,V_2\in \Omega^{(s)}$.

If for all $x\in X$,
 the $i$th element of  
$F^{(s)}_+ \circ \cdots \circ F^{(L)}_+ (x)$
  is zero in a neighborhood of
$$\{ A^{*(s)}, A^{*(s+1)},\ldots, A^{*(L)}, B^{*(s)}, B^{*(s+1)}, 
\ldots B^{*(L)}\},$$
then, we can remove the $i$th row of $A^{(s)}$, the $i$th element of $B^{(s)}$, and the $i$th column of $A^{(s-1)}$. This removal corresponds to eliminating the $i$th unit in the $s$th hidden layer.

{ 
\begin{example}
\label{example:relu}
Let 
${A^*}^{(1)}=\left(\begin{array}{ccccccc} 1&1
\end{array}\right)$,\\
${A^*}^{(2)}=\left(\begin{array}{ccccccc} 1&1\\-1&-1
\end{array}\right)$ and ${B^*}^{(2)}=\left(\begin{array}{c} -2\\-1
\end{array}\right)$.
Also let $x=\left(\begin{array}{c} x_1\\x_2\end{array}\right)$, $x_1,x_2\geq 0.$

Consider the case of 
$$y={A^*}^{(1)}({A^*}^{(2)}x+{B^*}^{(2)})_+= (1,1)
\left(\begin{array}{c} (x_1+x_2-2)_+ \\(-x_1-x_2-1)_+ \end{array}\right) .$$
We assume that $x_1$ and $x_2$ are positive, and
 therefore $-x_1-x_2-1$ is never positive. 
 So the $2$th row of $A^{(2)}$, the $2$th element of $B^{(2)}$, and the $2$th column of $A^{(1)}$
  can be removed.
\end{example}}

\begin{lemma}
\label{lemma:idealc}
Let $C_1,C_2,C_3,C_4,C_5$ be matrices
with  $C_5=C_1(C_2,C_3)C_4$.
Then we have
 $$
 \langle C_5,  C_1(C_2, O)C_4\rangle
= \langle C_1(C_2, O)C_4, C_1(O, C_3)C_4\rangle,$$
where $O$ is the zero matrix.
\end{lemma}

We have
$$K(w)=\int_{X}||h_+(x,A,B)-h_+(x,A^*,B^*)||^2q(x)dx$$
$$=\sum_{V\in \Omega^{(1)} :
{\rm Vol}(V)>0} \int_{V}||h_+(x,A,B)-h_+(x,A^*,B^*)||^2q(x)dx.$$
By Theorem \ref{theorem:k(w)}
and by Lemma \ref{lemma:idealc},
we have 
$$c_{w^*}(K(w))=c_{w^*}(\langle 
F^{(1)}_{i^{(1)}}\circ \cdots \circ F^{(L)}_{i^{(L)}}
\rangle),$$
where $i^{(s)}= \{i^{(s)}_1,i^{(s)}_2,\cdots,i^{(s)}_{k^{(s)}}\} \subset \{1,2,\cdots,H^{(s)}\}$,
$A_{i^{(s)}}=\left( \begin{array}{ccc}
a_{i^{(s)}_1,i^{(s+1)}_1}&\cdots& a_{i^{(s)}_1,i^{(s+1)}_{k^{(s+1)}}}\\
\vdots &\cdots& \vdots \\
a_{i^{(s)}_{k^{(s)}},i^{(s+1)}_1}&\cdots& a_{i^{(s)}_{k^{(s)}},i^{(s+1)}_{k^{(s+1)}}}
\end{array}\right)$
and

$B_{i^{(s)}}=\left( \begin{array}{ccc}
b_{i^{(s)}_1}\\
\vdots \\
b_{i^{(s)}_{k^{(s)}}}\end{array}\right)$.

In this paper, we consider the case in {\bf three}-layered neural networks with 
activation function ReLU (Rectified Linear Unit function).

\begin{theorem}
\label{theorem:theta}

For $\alpha = 1, \ldots, m$, 
let  $f^{(\alpha)}(u^{(\alpha)})$ and $g^{(\alpha)}(u^{(\alpha)})$
be analytic functions.
Also, let 
$$ c_{w^{*(\alpha)}}\left((f^{(\alpha)})^2, g^{(\alpha)}\right) = \lambda^{(\alpha)} \quad \text{and} \quad \theta_0\left((f^{(\alpha)})^2, g^{(\alpha)}\right) = \theta^{(\alpha)}. $$
Then we have
$$ c_{(w^{*(1)},\ldots,w^{*(m)})}\left(\langle f^{(1)}, \ldots, f^{(m)} \rangle, \prod_{\alpha=1}^m g^{(\alpha)}\right) = \sum_{\alpha=1}^m \lambda^{(\alpha)}, $$
and
$$ \theta_{(w^{*(1)},\ldots,w^{*(m)})}\left(\langle f^{(1)}, \ldots, f^{(m)} \rangle, \prod_{\alpha=1}^m g^{(\alpha)}\right) = \sum_{\alpha=1}^m (\theta^{(\alpha)} - 1) + 1. $$

\end{theorem}

\noindent
(Proof)

 Without loss of generality, we can assume that
$$ f^{(\alpha)} = (u^{(\alpha)}_1)^{k^{(\alpha)}_1} (u^{(\alpha)}_2)^{k^{(\alpha)}_2} \cdots (u^{(\alpha)}_{d^{(\alpha)}})^{k^{(\alpha)}_{d^{(\alpha)}}}, $$
$$ g^{(\alpha)} = (u^{(\alpha)}_1)^{h^{(\alpha)}_1} (u^{(\alpha)}_2)^{h^{(\alpha)}_2} \cdots (u^{(\alpha)}_{d^{(\alpha)}})^{h^{(\alpha)}_{d^{(\alpha)}}}. $$
Let $ u = (u^{(1)}_1, \ldots, u^{(1)}_{d^{(1)}}, \ldots, u^{(m)}_1, \ldots, u^{(m)}_{d^{(m)}}) $ and $ d = \sum_{\alpha=1}^m d^{(\alpha)} $.
Assume for $j \geq \theta^{(\alpha)} + 1$
$$ \lambda^{(\alpha)} = \frac{h^{(\alpha)}_1 + 1}{2k^{(\alpha)}_1} = \cdots = \frac{h^{(\alpha)}_{\theta^{(\alpha)}} + 1}{2k^{(\alpha)}_{\theta^{(\alpha)}}} 
< \frac{h^{(\alpha)}_j + 1}{2k^{(\alpha)}_j}. $$
By resolution of singularities, set  
$ u^{(\alpha)}_j = \tilde{u}_1^{\ell^{(\alpha)}_{1,j}} \cdots \tilde{u}_d^{\ell^{(\alpha)}_{d,j}}.$
For the matrix $L$ given by
$$
L = \begin{pmatrix}
\ell^{(1)}_{1,1} & \ell^{(1)}_{2,1} & \cdots & \ell^{(1)}_{d,1} \\
\vdots & & & \vdots \\
\ell^{(1)}_{1,d^{(1)}} & \ell^{(1)}_{2,d^{(1)}} & \cdots & \ell^{(1)}_{d,d^{(1)}} \\
\vdots & & & \vdots \\
\ell^{(m)}_{1,1} & \ell^{(m)}_{2,1} & \cdots & \ell^{(m)}_{d,1} \\
\vdots & & & \vdots \\
\ell^{(m)}_{1,d^{(m)}} & \ell^{(m)}_{2,d^{(m)}} & \cdots & \ell^{(m)}_{d,d^{(m)}}
\end{pmatrix},
$$
the determinant of Jacobian $ \left|\frac{\partial (u^{(1)}, \ldots, u^{(m)})}{\partial \tilde{u}}\right| $ is 
$$ \prod_{s=1}^d \tilde{u}_s^{-1 + \sum_{\alpha=1}^m \sum_{j=1}^{d^{(\alpha)}} \ell^{(\alpha)}_{s,j}} |L|. $$
We have
$ \prod_{\alpha=1}^m g^{(\alpha)} = \prod_{s=1}^d \tilde{u}_s^{\sum_{\alpha=1}^m \sum_{j=1}^{d^{(\alpha)}} \ell^{(\alpha)}_{s,j} h^{(\alpha)}_j}.$

Since $ {f^{(1)}}^2 + \cdots + {f^{(m)}}^2 $ is normal crossing, there exists $\alpha_0$ such that
$ {f^{(\alpha_0)}}^2 \geq {f^{(\alpha)}}^2 \quad (1 \leq \alpha \leq m). $
Therefore, we have
$$ \sum_{j=1}^{d^{(\alpha_0)}} \ell^{(\alpha_0)}_{s,j} 2k^{(\alpha_0)}_j \leq \sum_{j=1}^{d^{(\alpha)}} \ell^{(\alpha)}_{s,j} 2k^{(\alpha)}_j. $$
The possible candidates for the log canonical threshold are given by
\begin{eqnarray*}
&&\hspace*{-10mm}\frac{\sum_{\alpha=1}^m \sum_{j=1}^{d^{(\alpha)}} (\ell^{(\alpha)}_{s,j} h^{(\alpha)}_j + \ell^{(\alpha)}_{s,j})}{\sum_{j=1}^{d^{(\alpha_0)}} \ell^{(\alpha_0)}_{s,j} 2k^{(\alpha_0)}_j} \geq \sum_{\alpha=1}^m \frac{\sum_{j=1}^{d^{(\alpha)}} (\ell^{(\alpha)}_{s,j} h^{(\alpha)}_j + \ell^{(\alpha)}_{s,j})}{\sum_{j=1}^{d^{(\alpha)}} \ell^{(\alpha)}_{s,j} 2k^{(\alpha_0)}_j}\geq \sum_{\alpha=1}^m \lambda^{(\alpha)},
\end{eqnarray*}
where we use the inequality $\frac{s_3 + s_4}{s_1 + s_2} \geq \min\left\{ \frac{s_3}{s_1}, \frac{s_4}{s_2}\right\}$ for $ s_1, s_2, s_3, s_4 \geq 0 $.

On the other hand, to obtain 
$$ \sum_{\alpha=1}^m \lambda^{(\alpha)} = \frac{\sum_{\alpha=1}^m \sum_{j=1}^{d^{(\alpha)}} (\ell^{(\alpha)}_{s,j} h^{(\alpha)}_j + \ell^{(\alpha)}_{s,j})}{\sum_{j=1}^{d^{(\alpha_0)}} \ell^{(\alpha_0)}_{s,j} 2k^{(\alpha_0)}_j}, $$
requires  
$ \sum_{j=1}^{\theta^{(\alpha_0)}} \ell^{(\alpha_0)}_{s,j} 2k^{(\alpha_0)}_j = \sum_{j=1}^{\theta^{(\alpha)}} \ell^{(\alpha)}_{s,j} 2k^{(\alpha)}_j, $
and $\ell^{(\alpha)}_{s,j} = 0$ for $ j > \theta^{(\alpha)} $.
The dimension of the vectors satisfying such conditions is $ \sum_{\alpha=1}^m (\theta^{(\alpha)} - 1) + 1 $.

Finally, the Jacobian is nonzero if all vectors in $ L $ are linearly independent. These conditions complete the proof of the theorem.

\hfill{(Q.E.D.)}

Using Theorem \ref{theorem:theta},   we  have the following theorem.

\begin{theorem}

Let
\begin{eqnarray*}
h_{+}(x,A,B)&=&A^{(1)}(A^{(2)}x+B^{(2)})_{+}
\end{eqnarray*}

The model has 
$H^{(3)}$ input units, $H^{(1)}$ output units,
and $H^{(2)}$ hidden units in one hidden layer.
Let us divide the hidden layer into $k^{(2)}$ groups:
$$H^{(2)}_1+H^{(2)}_2+\cdots+H^{(2)}_{k^{(2)}}={H'}^{(2)},$$
{ where $H^{(2)} - {H'}^{(2)}$ represents the number of removed neurons.}
Let ${H'}^{(1)}_i$ be the number such that 
$H^{(1)} - {H'}^{(1)}_i$ represents the number of removed neurons for $1\leq i\leq k^{(2)}$, respectively.

Then, 
$\lambda$ and $\theta$ for the model 
corresponding to the log canonical threshold 
$$\lambda= \sum_{i=1 } ^{k^{(2)}}
\lambda({H'}^{(1)}_i ,H^{(2)}_i,H^{(3)}+1,r_i)$$
and
$$
\theta= \sum_{i=1 } ^{k^{(2)}}
(\theta({H'}^{(1)}_i ,H^{(2)}_i,H^{(3)}+1,r_i)-1)+1,$$
where $r_i$ is the rank corresponding the group $i$.

\end{theorem}

\section{Softmax function}

Next we consider the Softmax function.
Let $W$
be the set of parameters.
Denote the input value by $x\in {\bf R}^{H^{(L+1)}}$
with probability density function $q(x)$
and output value $y,z\in {\bf R}^{H^{(1)}}$
with $y=h(x,w)$ and 
$$z={\rm Softmax}(y|x,w)=( \frac{e^{y_1}}{\sum_{i}e^{y_i}},\cdots, \frac{e^{y_{H^{(L+1)}}}}{\sum_{i}e^{y_i}}).$$
Consider the  statistical model
with Gaussian noise,
$$p(y|x,w)=\frac{1}{(\sqrt{2\pi})^{H^{(1)}} }\exp
(-\frac{1}{2}||z-{\rm Softmax}(y|x,w)||^2). $$

Assume the set of optimal parameters $W_0$ by 
\begin{eqnarray*}
W_0&=&\{w_0\in W | L(w_0) =\min_{w'\in W} L(w')\}\\
&=&
\{ w \ | \  {\rm Softmax} (y|x,w)={\rm Softmax} (y|x,w_0)  \} ,
\end{eqnarray*}
and consider for $w_0\in W_0$,
$$c_{w_0}(|| {\rm Softmax} (y|x,w) - {\rm Softmax} (y|x,w_0)||^2).
$$

\begin{theorem}

We have
\begin{eqnarray*}
&&\langle  {\rm Softmax} (y|x,w) - {\rm Softmax} (y|x,w_0) \rangle\\
&&=\langle  {y_2(w)-y_1(w) -(y_2(w_0)-y_1(w_0))},
\cdots ,y_{H^{(1)}}(w)-y_1(w)-(y_{H^{(1)}}(w_0)-y_1(w_0)).\rangle 
\end{eqnarray*}
{  Therefore, $c_{w_0}(|| {\rm Softmax} (y|x,w) - {\rm Softmax} (y|x,w_0)||^2)$
$$=
c_{w_0}(|| \left(\begin{array}{c} h_2(x,w)-h_1(x,w)\\ h_3(x,w)-h_1(x,w)\\ \vdots \\h_{H^{(1)}}(x,w)-h_1(x,w)
\end{array}\right) -\left(\begin{array}{c} h_2(x,w_0)-h_1(x,w_0)\\ h_3(x,w_0)-h_1(x,w_0)\\ \vdots \\h_{H^{(1)}}(x,w_0)-h_1(x,w_0)
\end{array}\right)||^2)
.
$$
}

\end{theorem}

\noindent
(Proof)

Since  
$$\langle \frac{1}{f}-\frac{1}{f'}, \frac{g}{f}-\frac{g'}{f'} \rangle=\langle \frac{1}{f}-\frac{1}{f'}, \frac{g}{f}-\frac{g}{f'}+\frac{g}{f'}-\frac{g'}{f'} \rangle$$
$$=\langle \frac{1}{f}-\frac{1}{f'} , g(\frac{1}{f}-\frac{1}{f'})+\frac{g-g'}{f'} \rangle
=\langle \frac{1}{f}-\frac{1}{f'} , g-g' \rangle,$$
{ for $f,f',g,f' >0$,}
we obtain
\begin{eqnarray*}
&&\langle  {\rm Softmax} (y|x,w) - {\rm Softmax} (y|x,w_0) \rangle\\
&=&
\langle( \frac{e^{y_1(w)}}{\sum_{i}e^{y_i(w)}},\cdots, \frac{e^{y_{H^{(1)}(w)}}}{\sum_{i}e^{y_i(w)}})-
( \frac{e^{y_1(w_0)}}{\sum_{i}e^{y_i(w_0)}},\cdots, \frac{e^{y_{H^{(1)}}(w_0)}}{\sum_{i}e^{y_i(w_0)}})\rangle\\
&=&
\langle ( \frac{e^{y_1(w)-y_1(w)}}{\sum_{i}e^{y_i(w)-y_1(w)}},\cdots, \frac{e^{y_{H^{(1)}}(w)-y_1(w)}}{\sum_{i}e^{y_i(w)-y_1(w)}})-
( \frac{e^{y_1(w_0)-y_1(w_0)}}{\sum_{i}e^{y_i(w_0)-y_1(w_0)}},\cdots, \frac{e^{y_{H^{(1)}}(w_0)-y_1(w_0)}}{ \sum_{i}e^{y_i(w_0)-y_1(w_0)}})\rangle \\
&=&
\langle  \frac{1}{\sum_{i}e^{y_i(w)-y_1(w)}}-\frac{1}{\sum_{i}e^{y_i(w_0)-y_1(w_0)}},  \\
&&e^{y_2(w)-y_1(w)}-e^{y_2(w_0)-y_1(w_0)},\cdots, e^{y_{H^{(1)}}(w)-y_1(w)}-e^{y_{H^{(1)}}(w_0)-y_1(w_0)}\rangle \\
&=&
\langle  e^{y_2(w)-y_1(w)}-e^{y_2(w_0)-y_1(w_0)},
\cdots, e^{y_{H^{(1)}}(w)-y_1(w)}-e^{y_{H^{(1)}}(w_0)-y_1(w_0)}\rangle \\
&=&
\langle  e^{y_2(w)-y_1(w) -(y_2(w_0)-y_1(w_0))}-1,\cdots, e^{y_{H^{(1)}}(w)-y_1(w)-(y_{H^{(1)}}(w_0)-y_1(w_0))}-1\rangle \\
&=&
\langle  {y_2(w)-y_1(w) -(y_2(w_0)-y_1(w_0))},
\cdots, y_{H^{(1)}}(w)-y_1(w)-(y_{H^{(1)}}(w_0)-y_1(w_0))\rangle .
\end{eqnarray*}

\hfill{(Q.E.D.)}

{ 

\section{Conclusion}

In this paper, we studied the learning coefficients for deep neural networks with linear units and ReLU units.  
Throughout this study, we assume that the a priori probability density function \( \varphi(w) \) satisfies \( \varphi(w_0) > 0 \). This assumption is natural, as it ensures that the probability of having the optimal parameter \( w_0 \) is positive.  
If \( \varphi(w_0) = 0 \), then the learning model cannot reach the optimal parameter during training.

We showed that the learning coefficients for ReLU units are equivalent to those of neural networks with linear units by segmenting the input vector space.
This result underscores the effectiveness of ReLU units and builds upon a theoretical finding from \cite{AoMulti}.
 The learning coefficients \( \lambda \) decrease as the number of layers increases \cite{AoMulti}.  
Therefore, the Kullback-Leibler divergence between \( q((x,y)^n) \) and \( p((x,y)^n) \), given by  
\begin{eqnarray*}  
&&D(q((x,y)^n) \mid p((x,y)^n) )  
=\int q((x,y)^n) \log q((x,y)^n)  \prod_{i=1}^n dx_idy_i \\  
&&-\int q((x,y)^n) \log \prod_{i=1}^n p_0(x_i,y_i)  
\prod_{i=1}^n dx_idy_i  +\lambda(w_0) \log (n)-(\theta(w_0)-1)\log\log (n)+O_p(1),  
\end{eqnarray*}  
also decreases as the number of layers increases.  
This implies that while deeper models exhibit greater complexity, they also achieve improved effectiveness.
This is one of the theoretical reasons for the effectiveness of deep linear neural networks.

Once these theoretical values are established, they provide insights into the theoretical free energy and generalization error, 
which are essential for assessing probabilistic models.  
Moreover, these values serve as a benchmark for validating numerical computations.  
They have been effectively utilized in numerical experiments, including information criteria, Markov chain Monte Carlo methods ~\cite{Nagata1,Nagata2}, and model selection techniques.

}


\section*{Acknowledgements}

This research was funded by Grants-in-Aid for Scientific Research - KAKENHI - under grant number 24K15114.
We would like to express our gratitude to the reviewers for their insightful discussions, which greatly contributed to the enhancement of our paper.

\bibliographystyle{plain}
\bibliography{reference.bib}

\end{document}